\pdfoutput=1

\documentclass[11pt,a4paper]{article}
\usepackage[hyperref]{acl2023}
\usepackage{times}
\usepackage{inconsolata}
\usepackage{latexsym}
\usepackage{danudefs}
\usepackage{algorithmic}
\usepackage{algorithm}
\usepackage{color}
\usepackage{booktabs}
\usepackage{xspace}
\usepackage{url}
\usepackage{braket}
\usepackage{enumitem}
\usepackage{multicol}
\usepackage{multirow}
\usepackage{comment}
\usepackage{subcaption}

\usepackage{pifont}% http://ctan.org/pkg/pifont

\newcommand{\xmark}{\ding{55}}%

\usepackage[english]{babel}
\usepackage{hyperref}
\addto\captionsenglish{%
}
\addto\extrasenglish{%
}

\def\equationautorefname~#1\null{(#1)\null} % uncomment this line to get Equation~(1) formatiing (which can waste space!)
\usepackage{acronym}

\usepackage{etoolbox}
\makeatletter
\newif\if@in@acrolist
\AtBeginEnvironment{acronym}{\@in@acrolisttrue}
\newrobustcmd{\LU}[2]{\if@in@acrolist#1\else#2\fi}

\newcommand{\ACF}[1]{{\@in@acrolisttrue\acf{#1}}}
\makeatother

\acrodef{MLM}[MLM]{Masked Language Model}
\acrodefplural{MLM}[MLMs]{\LU{M}{m}asked \LU{L}{l}anguage \LU{M}{m}odels}
\acrodef{LLM}[LLM]{Large Language Model}
\acrodefplural{LLM}[LLMs]{Large Language Models}
\acrodef{SoTA}[SoTA]{state-of-the-art}
\acrodef{ICL}{\LU{I}{i}n-cotext \LU{L}{l}earning}
\acrodef{SCD}{\LU{S}{s}emantic \LU{C}{c}hange \LU{D}{d}etection}
\acrodef{WiC}{Word-in-Context}
\acrodef{ITML}[ITML]{Information-Theoretic Metric Learning}
\acrodef{SDML}[SDML]{Semantic Distance Metric Learning}
\acrodef{LSA}[LSA]{Latent Semantic Analysis}
\usepackage{todonotes}

\makeatletter
\if@todonotes@disabled

\else

\fi
\makeatother

\title{A Semantic Distance Metric Learning approach for \\Lexical Semantic Change Detection}
\author{Taichi Aida \\
  Tokyo Metropolitan University \\
  {\tt aida-taichi@ed.tmu.ac.jp}
  \And Danushka Bollegala \\
  University of Liverpool \\
  {\tt danushka@liverpool.ac.uk}}

\date{}

\begin{document}
\maketitle

\begin{abstract}
Detecting temporal semantic changes of words is an important task for various NLP applications that must make time-sensitive predictions.
Lexical \ac{SCD} task involves predicting whether a given target word, $w$, changes its meaning between two different text corpora, $C_1$ and $C_2$.
For this purpose, we propose a supervised two-staged \ac{SCD} method that uses existing \ac{WiC} datasets.
In the first stage, for a target word $w$, we learn two \emph{sense-aware} encoders that represent the meaning of $w$ in a given sentence selected from a corpus.
Next, in the second stage, we learn a \emph{sense-aware} distance metric that compares the semantic representations of a target word across all of its occurrences in $C_1$ and $C_2$.
%Experimental results on multiple benchmark datasets for \ac{SCD} show that our proposed method consistently outperforms all previously proposed \ac{SCD} methods for multiple languages, establishing a novel \ac{SoTA} for \ac{SCD}.\footnote{Source code is available at \url{https://github.com/a1da4/svp-sdml} .
Experimental results on multiple benchmark datasets for \ac{SCD} show that our proposed method achieves strong performance in multiple languages.
Additionally, our method achieves significant improvements on \ac{WiC} benchmarks compared to a sense-aware encoder with conventional distance functions.\footnote{Source code is available at \url{https://github.com/LivNLP/svp-sdml} .}
\end{abstract}

\section{Introduction}
\label{sec:intro}

The notion of word meaning is a dynamic one, and evolves over time as noted by~\newcite{tahmasebia-etal-2021-survey}.
For example, the meaning of the word \textit{cell} has changed over time to include \textit{cell phone} to its previous meanings of \textit{prison} and \textit{related to biology}.
Detection of such semantic changes of words over time remains a challenging, yet an important task for lexicography, sociology, and information retrieval~\citep{traugott-dasher-2001-regularity, cook-stevenson-2010-automatically, michel-etal-2011-quantitative, kutuzov-etal-2018-diachronic}.
For example, in E-commerce, a user might use the same keyword, such as \emph{scarf}, to search for different types of products based on seasonal variations, such as \emph{silk scarves in spring} versus \emph{woollen scarves in winter}. 
Recently, a decline in the performance of pretrained \acp{LLM} over time has been observed, and attributed to their training on static snapshots~\citep{loureiro-etal-2022-timelms, lazaridou-etal-2021-mind}. 
Recognising the words with changed meanings enables efficient fine-tuning of LLMs to incorporate only those with specific semantic shifts~\citep{Su-etal-2022-improving}.

\begin{figure}[t!]
    \centering
    \includegraphics[width=76mm]{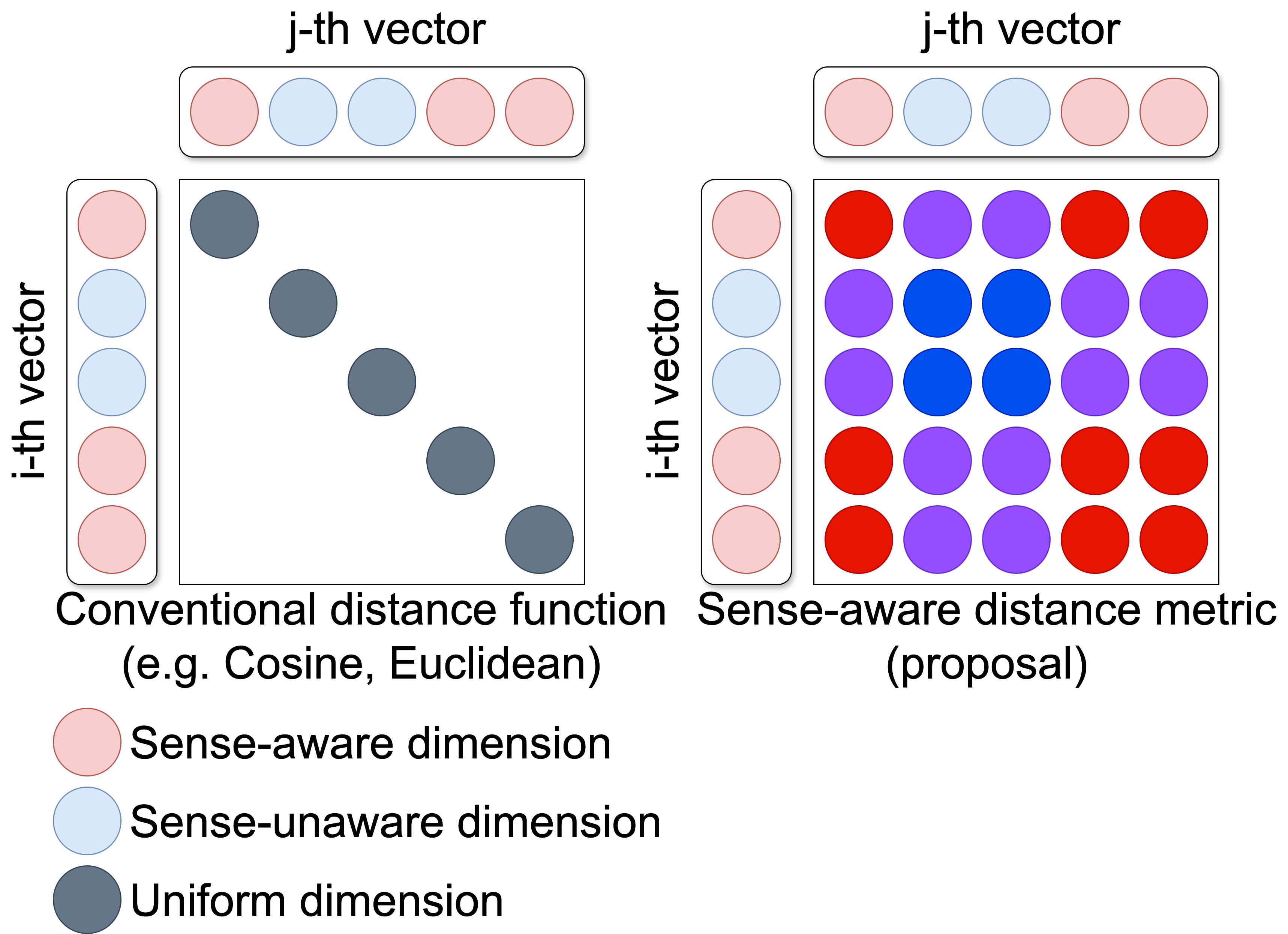}
    \caption{An overview of our method. Conventional distance functions such as Cosine or Euclidean consider the information of each dimension uniformly (left), but our method considers sense-aware information and cross-dimensional correlations (right).}
    \label{fig:overview}
\end{figure}

Detecting whether a word has its meaning changed between two given corpora, sampled at different points in time, requires overcoming two important challenges, which we name as the \emph{representational} and \emph{measurement} challenges.
\begin{description}[style=unboxed,leftmargin=0cm]
     \item [Representational Challenge:]
    A word can take different meanings in different contexts even within the same corpus. 
    Therefore, creating a representation for the meaning of a word across an entire corpus is a challenging task compared to that in a single sentence or a document.
    Prior work has averaged static~\citep{kim-etal-2014-temporal, kulkarni-etal-2015-statistically, hamilton-etal-2016-diachronic, yao-etal-2018-dynamic, dubossarsky-etal-2019-time, aida-etal-2021-comprehensive} or contextualised~\citep{martinc-etal-2020-leveraging, beck-2020-diasense, kutuzov-giulianelli-2020-uio, rosin-etal-2022-time, rosin-radinsky-2022-temporal} word embeddings for this purpose, which is suboptimal because averaging conflates multiple meanings of a word into a single vector. 

    \item [Measurement Challenge:]
    We require a distance metric that can accurately measure the semantic change of a target word between two given corpora, using the learnt representations of the target word from each corpus. 
    This challenge has been addressed in prior work using parameter-free distance functions due to the lack of labelled training data to provide any supervision for \ac{SCD} ~\citep{kutuzov-giulianelli-2020-uio, card-2023-substitution, cassotti-etal-2023-xl, aida-bollegala-2023-unsupervised, aida-bollegala-2023-swap, tang-etal-2023-word}.
    %While most conventional parameter-free distance functions treat all dimensions uniformly, \autoref{fig:analyze_dim_semeval_en} shows that the sentence embeddings have \emph{\ac{SCD}-aware} and \emph{\ac{SCD}-unaware} dimensions.
\end{description}

% Proposal: separate SCD method into two stages and use sense-aware supervision in both stages.
To address those challenges, we propose \ac{SDML}, a supervised two-staged \ac{SCD} method.
To solve the representational challenge, we learn a sense-aware\footnote{In this paper, the term `sense' refers not to strictly defined senses as found in dictionaries or WordNet~\cite{miller-1998-wordnet}, but rather to the nuances of word meanings derived from the distributional hypothesis~\cite{harris-1954-DS}.} encoder from sense-level supervision that represents the meaning of a target word in a sentence.
Word sense information has shown to be useful for \ac{SCD}~\citep{rachinskiy-and-arefyev-2021-glossreader, arefyev-etal-2021-deepmistake, cassotti-etal-2023-xl, tang-etal-2023-word} and outperform word representations that encode only time~\citep{rosin-etal-2022-time,rosin-radinsky-2022-temporal}.
Following \citet{cassotti-etal-2023-xl}, we use \ac{WiC}~\citep{pilehvar-camacho-collados-2019-wic} data, annotated for word sense discrimination for training the sense-aware encoder.

To solve the measurement challenge,  we propose a method to learn a sense-aware distance metric to compare two sense-aware embeddings for a target word in sentences selected from the two corpora.
Specifically, the distance metric is trained such that it returns a smaller value between two sentences where the target word takes the same meaning compared to that between two sentences which express different meanings of the target word.
We learn a distance metric that satisfies this criteria, using the weak-supervision provided by existing \ac{WiC} datasets.

Experimental results show that \ac{SDML} achieves exceptional performance in the \ac{SCD} task.
%\ac{SDML} achieves comparable or superior performance against strong baselines.
Remarkably, \ac{SDML} obtains performance improvements of 2-5\% over the current strong baselines.
While our focus is on the \ac{SCD} task, we also explore the applicability of our \ac{SDML} to the \ac{WiC} task.
In \ac{WiC} benchmarks, \ac{SDML} significantly enhances performance in multiple languages.
%\ac{SoTA} method, thereby establishing the new \ac{SoTA} in four \ac{SCD} benchmarks. 
These results demonstrate the effectiveness of learning both a sense-aware encoder and a sense-aware distance metric.
% analyze dimensions
%Moreover, our analysis shows that there exists \ac{SCD}-aware dimensions, unique to each language, in its sentence embedding space.
%This finding explains the superior performance reported by \ac{SDML} in \ac{WiC} and \ac{SCD} tasks.

\section{Related Work}
\label{sec:related}
% Overview / Introduction of SCD tasks
The diachronic semantic changes of words have been extensively investigated by linguists to explore how meanings evolve over time~\citep{traugott-dasher-2001-regularity}.
In recent years, the advent of diachronic corpora and advancements in representing word meanings have paved the way for automated \ac{SCD} in NLP~\citep{tahmasebia-etal-2021-survey}.
% SCD methods (unsupervised)
To detect semantically changed words in time-separated corpora, unsupervised \ac{SCD} methods compare word embeddings trained on the target corpora.
Many methods exist that align vector spaces over time spanned by static word embeddings, such as initialisation~\citep{kim-etal-2014-temporal}, alignment~\citep{kulkarni-etal-2015-statistically,hamilton-etal-2016-diachronic}, and joint learning~\citep{yao-etal-2018-dynamic,dubossarsky-etal-2019-time,aida-etal-2021-comprehensive}.
Likewise, methods for comparing sets of target word representations computed as contextualised word embeddings have also been proposed.
Such methods include comparing the average~\citep{martinc-etal-2020-leveraging,beck-2020-diasense,kutuzov-giulianelli-2020-uio,laicher-etal-2021-explaining,giulianelli-etal-2022-fire,rosin-etal-2022-time,rosin-radinsky-2022-temporal} or each pair of embeddings~\citep{kutuzov-giulianelli-2020-uio,laicher-etal-2021-explaining}.
Additionally, \citet{aida-bollegala-2023-unsupervised} and \citet{nagata-etal-2023-variance} have proposed methods that consider the variance in the sets of embeddings.
These unsupervised \ac{SCD} methods have been evaluated using unsupervised \ac{SCD} tasks~\citep{del-tredici-etal-2019-short,schlechtweg-etal-2020-semeval,kutuzov-pivovarova-2021-rushifteval}.

% SCD methods (supervised)
Due to the lack of manually-labelled data for lexical semantic change, prior work on \ac{SCD} have been limited to unsupervised approaches.
A notable exception is XL-LEXEME~\citep{cassotti-etal-2023-xl}, a supervised \ac{SCD} method that uses sense-level supervision.
They focused on the \ac{WiC} task~\citep{pilehvar-camacho-collados-2019-wic} designed to detect semantic differences of a target word in a given pair of sentences.
XL-LEXEME fine-tunes a pretrained multilingual \ac{MLM} across multilingual \ac{WiC} datasets, and achieves strong performance in some \ac{SCD} tasks. %current \ac{SoTA} performance in some \ac{SCD} tasks.
Recent work show \ac{SCD} is a challenging task for \ac{LLM}s such as GPT-3.5~\citep{sorensen-etal-2022-information,periti-etal-2024-chatgpt}, reporting low performance even with carefully designed prompts.

\section{Semantic Distance Metric Learning}
\label{sec:method}
% SDML

Given a target word $w$ and two corpora $C_1$ and $C_2$ sampled at distinct time points, the goal in \ac{SCD} is to predict a score for $w$ that indicates the degree of semantic change undergone by $w$ between $C_1$ and $C_2$.
For this purpose, we propose, \ac{SDML}, which overcomes the two challenges introduced in~\autoref{sec:intro}.
Specifically, to address the representational challenge, we first learn a sense-aware encoder in \autoref{sec:encoder}.
Next, to overcome the measurement challenge, we learn a sense-aware distance metric between two sense-aware representations of a target word computed from $C_1$ and $C_2$, as described in \autoref{sec:metric}.

\subsection{Learning Sense-Aware Encoder}
\label{sec:encoder}

Following prior work on \ac{SCD} that show contextualised word embeddings to outperform the static word embeddings~\citep{aida-bollegala-2023-unsupervised,rosin-radinsky-2022-temporal}, we represent the meaning of a word $w$ in a sentence $s$ by the $d$-dimensional token embedding,\footnote{When a word has been tokenised into multiple subtokens, we compute the average of the subtoken embeddings to create the token embedding} $\vec{f}(w,s;\theta) (\in \R^d)$, obtained from a pretrained \ac{MLM} parametrised by $\theta$.
Although contextualised word embeddings capture the contextual information of a word in a sentence~\citep{yi-zhou-2021-learning}, explicitly encoding word sense information has been shown to improve performance in \ac{SCD} tasks~\citep{tang-etal-2023-word}.

For this purpose, we follow \citet{cassotti-etal-2023-xl} and fine-tune the \ac{MLM} on \ac{WiC}.
\ac{WiC} contains tuples $(w, s_1, s_2, y)$, where the label $y = 1$ if $w$ has the same meaning in both sentences $s_1$ and $s_2$, and $y=0$ otherwise.
We use a Siamese bi-encoder approach~\citep{reimers-gurevych-2019-sentence}, which uses two encoders to produce respective embeddings for pairs of sentences. 
This architecture enables us to capture the semantic relationship between the pairs of sentences.
In this paper, we use the same \ac{MLM} encoder to obtain two representations, $\vec{w}_1 = \vec{f}(w, s_1; \theta)$ and $\vec{w}_2 = \vec{f}(w, s_2; \theta)$ for the meaning of $w$ in $s_1$ and $s_2$.
We then update $\theta$ such that the contrastive loss, $\ell_c$~\citep{Hadsell} given by \autoref{eq:contrastive-loss} is minimised.
\begin{align}
    \label{eq:contrastive-loss}
    \ell_c (\theta) = \frac{1}{2} \left( y\delta^2 + (1-y) \max(0, m-\delta)^2\right)
\end{align}
Here, we set the margin $m = 0.5$ and  $\delta = 1 - (\vec{w}_1\T\vec{w}_2)/\norm{\vec{w}_1}\norm{\vec{w}_2}$ is the cosine distance.
Note that the same encoder is used to compute both $\vec{w}_1$ and $\vec{w}_2$ here.
We use AdamW~\citep{AdamW} as the optimiser to minimise the contrastive loss in \autoref{eq:contrastive-loss} with the initial learning rate set to $10^{-5}$ and the weight decay coefficient set to 0.01.

\subsection{Learning Sense-Aware Distance Metrics}
\label{sec:metric}

Armed with the sense-aware encoder trained in \autoref{sec:encoder}, we are now ready to learn a sense-aware distance metric
$h(\vec{w}_1, \vec{w}_2; \mat{A})$ given by \autoref{eq:metric} that measures the semantic 
distance between two sense-aware embeddings $\vec{w}_1 = \vec{f}(w,s_1; \theta)$ and $\vec{w}_2 = \vec{f}(w,s_2; \theta)$, for $w$ in two sentences $s_1$ and $s_2$, respectively.
\begin{align}
    \label{eq:metric}
    h(\vec{w}_1, \vec{w}_2; \mat{A}) = (\vec{w}_1 - \vec{w}_2)\T \mat{A}  (\vec{w}_1 - \vec{w}_2)
\end{align}
Here, $\mat{A} \in \R^{d \times d}$ is a Mahalanobis matrix (assumed to be positive definite) that defines a (squared) distance metric, Mahalanobis distance.
Unlike conventional distance functions such as Euclidean distance, Mahalanobis distance weights each dimension according to its variance and accounts for cross-dimensional correlations.

We use the \ac{WiC} training data instances to learn $\mat{A}$ in \autoref{eq:metric}.
Specifically, we consider two types of constraints that must be satisfied by $h$ as follows.
For instances where $y=1$, the distance between $\vec{w}_1$ and $\vec{w}_2$ is required to be greater than a distance lower bound $\ell$ such that, $h(\vec{w}_1, \vec{w}_2) \geq \ell$.
Likewise, for instances where $y=0$, the distance between $\vec{w}_1$ and $\vec{w}_2$ is required to be lower than a distance upper bound $u$ such that, $h(\vec{w}_1, \vec{w}_2) \leq u$.
There exists a bijection (up to a scaling function) between the set of Mahalanobis distances and the set of equal mean (denoted by $\vec{\mu}$) multivariate Gaussian distributions given by $p(\vec{w}; \mat{A}) = \frac{1}{Z} \exp\left(-\frac{1}{2} h(\vec{w}, \vec{\mu}; \mat{A})\right)$, where $Z$ is a normalising constant and $\mat{A}\inv$ is the covariance of the distribution.
We can use this bijection to measure the distance between two Mahalanobis distance functions parametrised by positive definite matrices $\mat{A}_0$ and $\mat{A}$ using the relative entropy (KL-divergence) between the corresponding multivariate Gaussians, given by \autoref{eq:KL}.
\begin{align}
\small
    \label{eq:KL}
    \mathrm{KL}(p(\vec{w}; \mat{A}_0) \vert\vert p(\vec{w}; \mat{A})) = \int p(\vec{w}; \mat{A}_0) \log \frac{p(\vec{w}; \mat{A}_0)}{p(\vec{w}; \mat{A})} d\vec{w}
\end{align}
Relative entropy is a non-negative and convex function in $\mat{A}$.
The overall optimisation problem is then given by \autoref{eq:opt}.
\begin{align}
    \label{eq:opt}
		\min_{\mat{A}}  \quad & \mathrm{KL}(p(\vec{w}; \mat{A}_0) \vert\vert p(\vec{w}; \mat{A})) \nonumber \\
    \text{subject to} \quad & h(\vec{w}_1, \vec{w}_2) \leq u  \quad y=1, \\ \nonumber
                    & h(\vec{w}_1, \vec{w}_2) \geq \ell \quad y=0 .
\end{align}

The optimisation problem given in \autoref{eq:opt} can be expressed as a particular type of Bregman divergence, which can be efficiently solved using the Bregman's method~\citep{Censor:1997}.
We use \ac{ITML}~\citep{davis-etal-2007-itml} for learning a Mahalanobis matrix $\mat{A}$ that satisfies those requirements.
Further details of \ac{ITML} are described in \autoref{sec:appendix_ITML}.

\subsection{Measuring Temporal Semantic Change}
\label{sec:inference}

Using the sense-aware distance metric $h$ and the sense-aware encoder $f$ learnt as described in the previous sections, we can now compute the semantic change score, $\alpha(w, C_1, C_2)$, of $w$ between $C_1$ and $C_2$ as the average pairwise distance computed over $\cS_1(w)$ and $\cS_2(w)$, given by \autoref{eq:score}.
\begin{align}
    \small
    \label{eq:score}
    \frac{1}{n_1 n_2} \sum_{\substack{s_1 \in \cS_1(w) \\ s_2 \in \cS_2(w)}} h(\vec{f}(w,s_1;\theta), \vec{f}(w,s_2;\theta); \mat{A})
\end{align}
Here, we denote the set of sentences where $w$ occurs in $C_i$ to be $\cS_i(w)$ for $i = 1, 2$, and the corresponding number of sentences in each set to be $n_i = |\cS_i(w)|$. 
Unlike much prior work in \ac{SCD}, which first computes a single vector (often by averaging) for a target word $w$ from a corpus, thereby conflating its different meanings, \autoref{eq:score} computes representations per each occurrence of $w$ and compares the average over all distances. 
Although the total number of summations, $n_1 n_2$, in \autoref{eq:score} could become large for frequent $w$, semantic change scores can be efficiently computed by pre-computing and indexing the sense-aware embeddings for all occurrences of $w$ in each corpus.
Moreover, the summation in \autoref{eq:score} can be computed in parallel over different batches of sentences to obtain a highly efficient Map-Reduce version~\citep{MapReduce}.

\section{Experiments}
\label{sec:exp}

\subsection{Setting}
\label{subsec:setting}
% WiC -> XL-LEXEME -> SCD
To learn the sense-aware encoder and the distance metric described in \autoref{sec:method}, we use WiC datasets covering multiple languages as shown in \autoref{tab:data_wic}: XL-WiC~\citep{raganato-etal-2020-xlwic}, MCL-WiC~\citep{martelli-etal-2021-mclwic}, and AM$^2$iCo~\citep{liu-etal-2021-am2ico}.\footnote{XL-WiC and MCL-WiC are licensed under a Creative Commons Attribution-NonCommercial 4.0 License, and AM$^2$iCo is licensed under a Creative Commons Attribution 4.0 International Public license.}

\begin{table}[t]
    \centering
    \small{
    \begin{tabular}{llrrr} \toprule
        Dataset & Language & \#Train & \#Dev & \#Test \\ \midrule
        \multirow{3}{*}{XL-WiC}  & German   & 48k     & 8.9k  & 1.1k   \\
                & French   & 39k     & 8.6k  & 22k    \\
                & Italian  & 1.1k    & 0.2k  & 0.6k   \\ \midrule
        MCL-WiC & English  & 4.0k    & 0.5k  & 0.5k   \\ \midrule
        \multirow{10}{*}{AM$^2$iCo} & German & 50k & 0.5k & 1.0k \\
                & Russian  & 28k     & 0.5k  & 1.0k   \\
                & Japanese & 16k     & 0.5k  & 1.0k   \\
                & Chinese  & 13k     & 0.5k  & 1.0k   \\
                & Arabic   & 9.6k    & 0.5k  & 1.0k   \\
                & Korean   & 7.0k    & 0.5k  & 1.0k   \\
                & Finnish  & 6.3k    & 0.5k  & 1.0k   \\
                & Turkish  & 3.9k    & 0.5k  & 1.0k   \\
                & Indonesian & 1.6k  & 0.5k  & 1.0k   \\
                & Basque   & 1.0k    & 0.5k  & 1.0k   \\ \bottomrule
    \end{tabular}
    }
    \caption{Statistics of the \ac{WiC} datasets. \#Train, \#Dev, and \#Test shows the number of instances.}
    \label{tab:data_wic}
\end{table}

\begin{table}[t]
    \centering
    \small{
    \begin{tabular}{llcr} \toprule
        Dataset & Language & Time Period & \#Sentences \\ \midrule
        \multirow{8}{*}{SemEval} & \multirow{2}{*}{English} & 1810--1860 & 254k  \\
         & & 1960--2010 & 354k \\
         & \multirow{2}{*}{German} & 1800--1899 & 2.6M \\
         & & 1946--1990 & 3.5M \\
         & \multirow{2}{*}{Swedish} & 1790--1830 & 3.4M \\
         & & 1895--1903 & 5.2M \\
         & \multirow{2}{*}{Latin} & B.C. 200--0 & 96k  \\
         & & 0--2000 & 463k \\ \midrule
        \multirow{3}{*}{RuShiftEval} & \multirow{3}{*}{Russian} & 1700--1916 & 3.3k \\
         & & 1918--1990 & 3.3k \\
         & & 1992--2016 & 3.3k \\ \bottomrule
    \end{tabular}
    }
    \caption{Statistics of the \ac{SCD} datasets. In the RuShiftEval, we used annotated pairwise sentences for prediction as in \citet{cassotti-etal-2023-xl}, and evaluation is conducted in three subsets: pre-Soviet~(1700--1916) vs. Soviet~(1918--1990), Soviet~(1918--1990) vs. post-Soviet~(1992--2016), and pre-Soviet~(1700--1916) vs. post-Soviet~(1992--2016), respectively referred to as RuShiftEval1 (Ru$_1$), RuShiftEval2 (Ru$_2$), and RuShiftEval3 (Ru$_3$). Full data statistics are shown in \autoref{sec:appendix_data}.}
    \label{tab:data_scd}
\end{table}

We use the sense-aware encoder released by \citet{cassotti-etal-2023-xl} as $f$,\footnote{This model is available at \url{https://huggingface.co/pierluigic/xl-lexeme}} which is based on XLM-RoBERTa$_\mathrm{large}$~\citep{conneau-etal-2020-xlmroberta} for the remainder of the experiments with \ac{SDML} reported in this paper.
We use the \texttt{metric\_learn}\footnote{\url{https://github.com/scikit-learn-contrib/metric-learn}} package to train a sense-aware distance metric with \ac{ITML}.
%It was trained on the \ac{WiC} datasets training data for each language using sense-aware contextualised embeddings obtained from XL-LEXEME.
The slack parameter $\gamma$ in \ac{ITML} weights the margin violations, and is searched from the 11 values in $\{10^{-5}, 10^{-4}, ..., 10^{0}, ..., 10^{4}, 10^{5}\}$, according to the best performance measured on the development data in each \ac{WiC} dataset.

\begin{table*}[t]
    \centering
    \small{
    \begin{tabular}{lcccccc} \toprule
    \multirow{2}{*}{Models} & MCL-WiC & \multicolumn{3}{c}{XL-WiC} & \multicolumn{2}{c}{AM$^2$iCo} \\
     & En & De & Fr & It & De & Ru \\ \midrule
    %\textbf{Baseline} & 84.0 & 76.2 & 72.3 \\
    \textbf{Baseline: Sense-aware Sentence Encoder} & 78.0 & 78.3 & 73.2 & 67.1 & 78.1 & 78.2 \\
     + Sense-aware Distance Metric & \textbf{90.3}$^{\dagger\dagger}$ & \textbf{84.9}$^{\dagger\dagger}$ & \textbf{78.7}$^{\dagger\dagger}$ & \textbf{75.3}$^{\dagger\dagger}$ & \textbf{85.0}$^{\dagger\dagger}$ & \textbf{87.6}$^{\dagger\dagger}$ \\ \bottomrule
    \end{tabular}
    }
    \caption{Accuracy of the \ac{WiC} test sets for languages relevant to the \ac{SCD} benchmarks. $\dagger\dagger$ indicates significance at the 95\% confidence interval.}
    \label{tab:result_wic_relevant_scd}
\end{table*}

After that, we evaluate the performance on \ac{SCD} tasks.
In this paper, we use the two benchmark datasets -- SemEval-2020 Task 1~\citep{schlechtweg-etal-2020-semeval} (covering English (En), German (De), Swedish (Sv) and Latin (La)) and RuShiftEval~\citep{kutuzov-pivovarova-2021-rushifteval} (covering Russian (Ru)), which have also been used in prior work on \ac{SCD}~\citep{kutuzov-etal-2021-grammatical, giulianelli-etal-2022-fire, cassotti-etal-2023-xl}.\footnote{SemEval-2020 Task 1 and RuShiftEval are licensed under a Creative Commons Attribution 4.0
International License and a GNU General Public License version 3.0, respectively.}
Statistics of those datasets are summarised in \autoref{tab:data_scd}.

For English, German and Russian, there exist \ac{WiC} training data splits that we can use to train \ac{SDML} separately for each of those languages.
However, no such training data are available for Latin and Swedish languages.
Therefore, when evaluating \ac{SDML} for Latin, we train it on the \ac{WiC} training data available for Italian and French, which are in the same Romance language family.
Likewise, when evaluating \ac{SDML} for Swedish, we train it on the \ac{WiC} training data available for German, which is in the same Germanic language family.
Creating \ac{WiC} datasets for languages that do not have such resources is beyond the scope of this paper and is deferred to future work.

Before presenting the results of the \ac{SCD} task, we first look at the performance of our \ac{SDML} on the \ac{WiC} task for languages corresponding to the \ac{SCD} benchmarks.
As a baseline, we use a sense-aware encoder fine-tuned to distinguish the different meanings of a target word by optimising~\eqref{eq:contrastive-loss}.
For \ac{SDML}, the sense-aware distance metric is learned via \ac{ITML} using \ac{WiC} datasets, classification boundaries are also obtained.
Therefore, we can use these boundaries for the prediction of the \ac{WiC} tasks.

\autoref{tab:result_wic_relevant_scd} shows that the combination of the sense-aware encoder and the sense-aware distance metric (our \ac{SDML}) constantly outperforms the baseline.
Interestingly, \ac{SDML} achieves significant improvements at the 95\% confidence interval computed using Bernoulli trials in all languages.
These results support our hypothesis: even in the task of detecting semantic differences at the same time period, it is better to use the combination of the sense-aware encoder and the sense-aware distance metric than the sense-aware encoder only.
More comprehensive results can be found in \autoref{subsec:results_wic}.

\subsection{Evaluating Semantic Changes of Words}
\label{subsec:results_scd}

\begin{table*}[t!]
    \centering
    \small{
    \begin{tabular}{lcccccccc} \toprule
        \multirow{2}{*}{Models} & \multirow{2}{*}{FT} & \multicolumn{4}{c}{SemEval} & \multicolumn{3}{c}{RuShiftEval} \\
         & & En & De & Sv & La & Ru$_1$ & Ru$_2$ & Ru$_3$ \\ \midrule
         \multicolumn{9}{l}{\textbf{Baselines: \ac{MLM}}} \\
          + APD~\citep{laicher-etal-2021-explaining} &            & 0.571 & 0.407 & 0.554 & -     & - & - & - \\
          + SSCD~\citep{aida-bollegala-2023-swap}   &            & 0.383 & 0.597 & 0.234 & 0.433 & - & - & - \\
          + APD~\citep{kutuzov-giulianelli-2020-uio} & \checkmark & 0.605 & 0.560 & 0.569 & 0.113 & - & - & - \\
          + ScaledJSD~\citep{card-2023-substitution} & \checkmark & 0.547 & 0.563 & 0.310 & 0.533 & - & - & - \\
          w/ TA + CD~\citep{rosin-radinsky-2022-temporal} & \checkmark & 0.520 & 0.763 & -     & 0.565 & - & - & - \\ 
          + CD \& UD + CD~\citep{giulianelli-etal-2022-fire} & \checkmark & 0.451 & 0.354 & 0.356 & \textbf{0.572} & 0.117 & 0.269 & 0.326 \\
         \midrule
         \multicolumn{9}{l}{\textbf{Sense-Aware Methods}} \\
         DeepMistake~\citep{arefyev-etal-2021-deepmistake} & \checkmark & - & - & - & - & 0.798 & 0.773 & 0.803 \\
         GlossReader~\citep{rachinskiy-and-arefyev-2021-glossreader} & \checkmark & - & - & - & - & 0.781 & 0.803 & 0.822 \\
         XL-LEXEME~\citep{cassotti-etal-2023-xl} &            & 0.757 & 0.877 & \textbf{0.754} & 0.056 & 0.775 & 0.822 & 0.809  \\ 
         XL-LEXEME~\citep{cassotti-etal-2023-xl} & \checkmark & - & - & - & - & 0.799 & \textbf{0.833} & 0.842 \\ %\midrule
         %\multicolumn{9}{c}{Proposal: SDML} \\ \midrule
         SDML (ours)            &            & \textbf{0.774} & \textbf{0.902} & 0.656 & 0.124 & \textbf{0.805} & 0.811 & \textbf{0.846} \\ \bottomrule
    \end{tabular}
    }
    \caption{Spearman's rank correlation on \ac{SCD} tasks compared against strong baselines. FT indicates whether fine-tuning on target time separated corpora was conducted. The absolute correlations for previous methods are taken from the respective papers as reported. - indicates that the corresponding benchmark was not evaluated in the referenced paper.}
    \label{tab:result_scd}
\end{table*}

We compare the performance of \ac{SDML} against prior work on several \ac{SCD} benchmarks.
In this evaluation, given a set of target words, an \ac{SCD} method under evaluation is required to predict scores that indicate the degree of semantic changes undergone by each word in the set.
The Spearman's rank correlation coefficient $r$ ($ \in [-1,1]$) between those predicted semantic change scores and that assigned by the human annotators in each benchmark dataset (i.e. \emph{gold} ratings) is computed as the evaluation metric.
An \ac{SCD} method that reports a high Spearman's $r$ value indicates better agreement with human ratings, and is considered to be desirable for detecting the semantic changes of words over time.

We compare the proposed methods against strong baselines %including \ac{SoTA} methods on the benchmarks 
as described next.
\paragraph{Baselines:}
\citet{kutuzov-giulianelli-2020-uio} and \citet{laicher-etal-2021-explaining} showed that the average pairwise cosine distance of the sets of contextualised embeddings perform well (\textbf{APD}).
\citet{aida-bollegala-2023-swap} proposed swapping-based \ac{SCD}, conducting context-swapping across target corpora to obtain more reliable scores (\textbf{SSCD}).
\citet{card-2023-substitution} proposed a token replacement-based JS divergence metric, to mitigate the influence of token frequency by replacing tokens with neighbouring words (\textbf{ScaledJSD}).
\citet{rosin-radinsky-2022-temporal} proposed temporal attention,  additional time-aware attention mechanism to \ac{MLM}s. 
SCD scores are calculated by the average cosine distance (\textbf{w/~TA~+~CD}).
\citet{giulianelli-etal-2022-fire} introduced an ensemble method combining the average cosine distance from \ac{MLM}s with the cosine distance between morpho-syntactic features labelled from universal dependencies (\textbf{CD~\&~UD~+~CD}). 
%It is the current \ac{SoTA} method in SemEval La.

\paragraph{Sense-Aware Methods:}
Here, we introduce sense-aware methods that use sense-level supervision.
\textbf{DeepMistake}~\citep{arefyev-etal-2021-deepmistake} and \textbf{GlossReader}~\citep{rachinskiy-and-arefyev-2021-glossreader} both leverage data from the \ac{WiC} and Word Sense Disambiguation tasks to fine-tune \acp{MLM}. 
Subsequently, both methods make predictions using linear regression trained on the provided training data. 
\textbf{XL-LEXEME} %, the current \ac{SoTA} method in SemEval En, De, SV, and RuShiftEval1-3, 
is fine-tuned on \ac{WiC} data across multiple languages. 
In our method, \ac{SCD} is performed using the average pairwise distance.
For \ac{SDML}, we predict the semantic change scores for the target words using \autoref{eq:score}.

Results are shown in \autoref{tab:result_scd}.\footnote{In SemEval De, Sv, and La, we have several models due to multiple \ac{WiC} datasets in the corresponding/related languages, but we report the highest performance.}
Our method performs comparable or superior to previous sense-aware methods using a sense-aware encoder only.
%We can see that our method achieves \ac{SoTA} performance in four benchmarks: SemEval En, De, RushiftEval-1 (Ru$_1$), and RushiftEval-3 (Ru$_3$).
Moreover, \ac{SDML} accomplishes performance improvements of 2-5\% over the %current \ac{SoTA} methods
sense-aware methods\footnote{In \ac{SCD} tasks, there is no statistical significance across ALL methods due to the lack of target words when we use the Fisher transformation.}, provided that training data in the corresponding languages is available.
In SemEval Sv and La, no training data for the corresponding languages exist, but our method performed equally well for SemEval Sv and showed a drastic performance improvement for SemEval La.

However, the performance in Latin is lower than pretrained MLMs in \autoref{tab:result_scd}. 
We attribute these results to the composition of the data used during MLM pretraining and SDML training, respectively.
Firstly, while Latin is included in the pretraining of XLM-RoBERTa, the dataset for pretraining is only a fraction (1/10 to 1/100) of the size of the datasets for the other languages evaluated in the \ac{SCD} benchmarks.
Secondly, Latin is also absent from the \ac{WiC} dataset used to train our \ac{SDML}. 
Therefore, we believe that the poor information on Latin, which was scarce even during the pretraining phase, was further diluted by the training of the \ac{SDML}, resulting in a low performance.
In the future, as the \ac{WiC} datasets are expanded and include more languages, it is expected that the performance of models for a wider range of languages will improve.

\subsection{Applying to \ac{WiC} Prediction}
%\subsection{Detecting Semantic Differences within the \emph{Same} Time Period (\ac{WiC})}
\label{subsec:results_wic}

% Explain why we do this task. This is not SCD. We still do this to check whether SDML can detect sense changes of words in two sentences (not over two corpora). 
% May be the results are SoTA on WiC benchmarks as well? Check this.

% train がある言語だけ表示
\begin{table*}[t]
    \small
    \centering
    \begin{tabular}{lcccccccccc} \toprule
        \multirow{2}{*}{Models} & MCL-WiC & \multicolumn{3}{c}{XL-WiC all} & \multicolumn{3}{c}{XL-WiC IV} & \multicolumn{3}{c}{XL-WiC OOV} \\
         & En & De & Fr & It & De & Fr & It & De & Fr & It \\ \midrule
         \multicolumn{11}{l}{\textbf{Baselines: MLM}} \\
         mBERT$_\mathrm{base}$ & 84.0 & 81.6 & 73.7 & 72.0 & 81.9 & 72.9 & 73.2 & 70.1 & 71.2 & 68.5 \\ 
         XLM-RoBERTa$_\mathrm{base}$ & 86.6 & 80.8 & 73.1 & 68.6 & 81.2 & 71.9 & 70.7 & 71.3 & 71.1 & 62.4 \\
         XLM-RoBERTa$_\mathrm{large}$ & - & 84.0 & 76.2 & 72.3 & 82.2 & 75.6 & 75.1 & 72.5 & 73.9 & 65.2 \\
         Lang. BERT & - & 82.9 & 78.1 & 72.6 & 83.2 & 77.6 & 73.9 & \textbf{76.6} & \textbf{78.0} & 69.1 \\ \midrule
         \multicolumn{11}{l}{\textbf{Sense-Aware Methods}} \\
         XL-LEXEME & 78.0 & 78.3 & 73.2 & 67.1 & 78.4 & 73.6 & 71.2 & 65.2 & 65.7 & 57.9 \\ %\midrule
         %\multicolumn{11}{c}{Proposal: SDML} \\ \midrule
         SDML (ours) & \textbf{90.3}$^\dagger$ & \textbf{84.9}$^\dagger$ & \textbf{78.7}$^{\dagger\dagger}$ & \textbf{75.3} & \textbf{85.1}$^\dagger$ & \textbf{78.5}$^{\dagger\dagger}$ & \textbf{77.8} & \textbf{76.6} & 75.3 & \textbf{70.2} \\ \bottomrule
    \end{tabular}
    \caption{Accuracy reported by different methods on the MCL-WiC and XL-WiC datasets. XL-WiC contains additional two test sets; in-vocabulary (IV) and out-of-vocabulary (OOV) test sets. Lang. BERT means language-specific BERT models. $\dagger$ or $\dagger\dagger$ indicate significance at the 90\% or 95\% confidence interval, respectively.}
    \label{tab:result_xlwic_mclwic}
\end{table*}

% train がある言語だけ表示
\begin{table*}[t]
    \small
    \centering
    \begin{tabular}{lcccccccccc} \toprule
        Models & De & Ru & Ja & Zh & Ar & Ko & Fi & Tr & Id & Eu \\ \midrule
        \multicolumn{11}{l}{\textbf{Baselines: MLM}} \\
        mBERT$_\mathrm{base}$ & 80.4 & 82.1 & 78.2 & 75.2 & 73.3 & 75.8 & 81.2 & 80.6 & 78.4 & 75.9 \\ 
        XLM-RoBERTa$_\mathrm{base}$ & 79.4 & 80.9 & 79.4 & 76.1 & 73.6 & 76.0 & 81.2 & 80.5 & 77.9 & 74.2 \\ \midrule
        \multicolumn{11}{l}{\textbf{Sense-Aware Methods}} \\
        XL-LEXEME & 78.1 & 78.2 & 77.1 & 75.2 & 75.4 & 75.5 & 78.0 & 78.7 & 75.5 & 72.7 \\ %\midrule
        %\multicolumn{11}{c}{Proposal: SDML} \\ \midrule
        SDML (ours) & \textbf{85.0}$^\dagger$ & \textbf{87.6}$^{\dagger\dagger}$ & \textbf{82.9} & \textbf{81.7}$^{\dagger\dagger}$ & \textbf{81.8}$^{\dagger\dagger}$ & \textbf{82.9}$^{\dagger\dagger}$ & \textbf{87.7}$^{\dagger\dagger}$ & \textbf{84.4} & \textbf{83.6}$^{\dagger\dagger}$ & \textbf{80.8}$^{\dagger}$ \\ \bottomrule
    \end{tabular}
    \caption{Accuracy of AM$^2$iCo dataset. $\dagger$ or $\dagger\dagger$ indicate significance at the 90\% or 95\% confidence interval, respectively.}
    \label{tab:result_am2ico}
\end{table*}

Our main focus in this paper has been Lexical Semantic Change Detection -- detecting whether a target word, $w$, has its meaning changed from one corpus, $C_1$, to another, $C_2$.
However, we use \ac{WiC} datasets for training the sense-aware encoder (described in \autoref{sec:encoder}) as well as the sense-aware distance metric (described in \autoref{sec:metric}) because there does not exist sufficiently large manually annotated datasets for training temporal \ac{SCD} methods.
Although none of the \ac{WiC} datasets we used are sampled from temporally distinct corpora, they provide a convenient alternative for training models that discriminate the meaning of a word in two given sentences.
Therefore, by evaluating our proposed \ac{SDML} on benchmark datasets for \ac{WiC}, we will be able to sanity check whether \ac{SDML} can indeed detect words that have same/different meanings in two given sentences, even if the two sentences might not be sampled from temporally distinct corpora.
In this \ac{WiC} task, a model is required to predict whether a target word takes the same meaning in two given sentences.
This is modelled as a binary classification task and classification accuracy is used as the evaluation metric.
A random prediction baseline would report an accuracy of $50\%$ on \ac{WiC} test datasets, which are balanced.

\paragraph{Baselines:}
As baselines for comparison, use report the accuracy of binary classifiers that have been trained using different \ac{MLM}s for the \ac{WiC} task as follows.
Given an instance $(w,s_1, s_2, y)$ from a \ac{WiC} training dataset, the token embeddings $\vec{f}(w,s_1)$ and $\vec{f}(w,s_2)$ are concatenated to represent the meaning of $w$ in $s_1$ and $s_2$.
Next, a binary logistic regression classifier is trained on the training data from a \ac{WiC} dataset. 
Note that the parameters of the \ac{MLM}s are not updated during this process.
For these baselines, we use the multilingual \ac{MLM}s such as mBERT$_\mathrm{base}$ and XLM-RoBERTa$_{\mathrm{base}/\mathrm{large}}$ as well as target language-specific BERT models (Lang. BERT) to train the baselines in XL-WiC such as BERT-base-german-cased\footnote{\url{https://huggingface.co/dbmdz/bert-base-german-cased}} for German, CamemBERT-large\footnote{\url{https://huggingface.co/almanach/camembert-large}} for French, and BERT-base-italian-xxl-cased\footnote{\url{https://huggingface.co/dbmdz/bert-base-italian-xxl-cased}} for Italian.
Rather than re-implementing or re-running these methods, we use the results presented in the corresponding benchmarks.

\paragraph{Sense-Aware Methods:}
Unlike the \ac{MLM}s used in the above-mentioned baselines, XL-LEXEME uses a sense-aware encoder fine-tuned to discriminate the different meanings of a target word.
XL-LEXEME is already fine-tuned using \ac{WiC} datasets and does not require a binary classifier to be trained, and predicts cosine distances less than the margin $m$ (set to 0.5) to be instances where the target word takes the same meaning in both sentences.
For \ac{SDML}, we use the sense-aware distance metric learned via \ac{ITML} to predict whether a target word expresses the same meaning in the two given sentences in \ac{WiC} datasets.
Since \ac{ITML} also learns classification boundaries (i.e. the upper and lower bounds denoted by respectively $u$ and $l$ in \autoref{algo:ITML}), which can be used to make binary predictions. 
Note that however, we do not have two different corpora in this \ac{WiC}-based evaluation, thus not requiring to average the distances as done in \autoref{eq:score}, and instead can perform a single point estimate considering the two sentences in a single \ac{WiC} test instance.

\autoref{tab:result_xlwic_mclwic} shows the results for MCL-WiC and XL-WiC datasets.
XL-WiC has in-vocab (IV) test sets and out-of-vocab (OOV) test sets extracted from the test set, in addition to the vanilla test set.
The results show that our \ac{SDML} consistently outperforms all of the baselines.
It shows 90\% accuracy in English, which is a significant improvement compared to XL-LEXEME, using the sense-aware distance metric.
In German and French, \ac{SDML} outperforms XLM-RoBERTa$_\mathrm{large}$, which is the same model as XL-LEXEME, as well as language-specific language models (Lang. BERT).
Interestingly, even in Italian, where there are only a few thousand training data instances~(\autoref{tab:data_wic}), the performance was significantly improved from XL-LEXEME.
Moreover, statistical tests reveal that our \ac{SDML} achieves significant improvements over all other methods on multiple test sets, with statistical significance observed at the 90\% or 95\% for binomial proportion confidence intervals.
%This shows the importance of using both a sense-aware sentence encoder and a sense-aware distance metric.

Similarly, in the AM$^2$iCo dataset, \autoref{tab:result_am2ico} also demonstrates the performance improvements achieved by \ac{SDML}.
%\autoref{tab:result_am2ico} shows the results for AM2iCo dataset.
In particular, according to \autoref{tab:result_am2ico}, \ac{SDML} achieves the best performance on all test sets.
Statistical tests demonstrate the significant performance enhancements of our \ac{SDML} compared to the other methods on most test sets, with significance observed at the 90\% or 95\% for binomial proportional confidence intervals.
While regular multilingual \ac{MLM}s and XL-LEXEME do not perform well for low-resource language (Eu) or languages not similar to English (Ja, Zh, Ar, and Ko), sense-aware distance metrics can dramatically improve performance for even just a few thousand instances.

\begin{table*}[t!]
    \centering
    \small{
    \begin{tabular}{lccccccc} \toprule
        \multirow{2}{*}{Models} & \multicolumn{4}{c}{SemEval} & \multicolumn{3}{c}{RuShiftEval} \\
         & En & De & Sv & La & Ru$_1$ & Ru$_2$ & Ru$_3$ \\ \midrule
         \textbf{Baseline: Sense-aware Sentence Encoder} & 0.757 & 0.877 & \textbf{0.754} & 0.056 & 0.775 & \textbf{0.822} & 0.809  \\ 
          + Sense-aware Distance Metric (diagonal) & 0.750 & \textbf{0.902} & 0.642 & 0.083 & 0.804 & 0.808 & \textbf{0.846} \\
          + Sense-aware Distance Metric (full) & \textbf{0.774} & \textbf{0.902} & 0.656 & \textbf{0.124} & \textbf{0.805} & 0.811 & \textbf{0.846} \\ \bottomrule
    \end{tabular}
    }
    \caption{Spearman's rank correlation on \ac{SCD} tasks.}
    \label{tab:ablation_scd}
\end{table*}

\begin{table*}[t!]
\small
    \centering
    %\begin{tabular}{l|p{3mm}p{3mm}|p{5mm}p{10mm}p{8mm}} \toprule
    \begin{tabular}{l|rc|rrr|rrr} \toprule
        \multirow{2}{1em}{Word} & \multicolumn{2}{c|}{Gold} & \multicolumn{1}{c}{WordNet} & \multicolumn{2}{c|}{Frequency} & Baseline & SDML$_\textit{diag}$ & SDML$_{\mathrm{full}}$ \\
         & rank & $\Delta$ & \#Synsets & \multicolumn{1}{c}{$C_1$} & \multicolumn{1}{c|}{$C_2$}  & \multicolumn{1}{c}{rank} & \multicolumn{1}{c}{rank} & \multicolumn{1}{c}{rank} \\ \midrule
        plane & 1 & \checkmark & 5 & 278 & 792 & 1 & 1 & 1 \\
        tip & 2 & \checkmark & 9 & 119 & 241 & 3 & 6 & 4 \\
        prop & 3 & \checkmark & 3 & 121 & 147 & 9 & 7 & 7 \\ 
        graft & 4 & \checkmark & 3 & 119 & 109 & 7 & 2 & 6 \\
        record & 5 & \checkmark & 8 & 420 & 1188 & 5 & 4 & 2 \\
        stab & 7 & \checkmark & 3 & 92 & 117 & 4 & 8 & 9 \\
        bit & 9 & \checkmark & 11 & 296 & 622 & 13 & 14 & 13 \\ 
        head & 10 & \checkmark & 33 & 3599 & 4127 & 14 & 18 & 14 \\
        \midrule
        multitude & 30 & \xmark & 3 & 475 & 131 & 27 & 26 & 24 \\
        savage & 31 & \xmark & 2 & 504 & 133 & 15 & 13 & 17 \\
        contemplation & 32 & \xmark & 2 & 240 & 111 & 28 & 27 & 28 \\ 
        tree & 33 & \xmark & 3 & 2322 & 1596 & 32 & 32 & 35 \\
        relationship & 34 & \xmark & 4 & 130 & 841 & 24 & 23 & 27 \\
        fiction & 35 & \xmark & 2 & 202 & 326 & 35 & 33 & 33 \\
        chairman & 36 & \xmark & 1 & 147 & 683 & 36 & 36 & 34 \\
        risk & 37 & \xmark & 4 & 286 & 643 & 37 & 37 & 37 \\
        \midrule
        \midrule
        Spearman & \multicolumn{2}{c|}{1.000} & 0.420 & $-$0.153 & $-$0.047 & 0.757 & 0.750 & \textbf{0.774} \\
        \bottomrule
    \end{tabular}
    \caption{Ablation study on the words categorised by the existence of semantic change: highlighting the top-8 semantically changed words with significant semantic change ($\Delta=$ \checkmark) and bottom-8 stable words with minimal semantic change ($\Delta=$ \xmark) on SemEval En. Baseline is a sense-aware sentence encoder only. \#Synsets shows the number of synsets in WordNet.\footnotemark}
    \label{tab:ablation_analysis}
\end{table*}
\footnotetext{\url{http://wordnetweb.princeton.edu/perl/webwn}}

\begin{table}[t!]
\small
    \centering
    \begin{tabular}{l|rrrr} \toprule
        \multirow{2}{*}{Models} & \multirow{2}{*}{Gold} & WordNet & \multicolumn{2}{c}{Frequency} \\ 
        & & \#Synsets & \multicolumn{1}{c}{$C_1$} & \multicolumn{1}{c}{$C_2$} \\ \midrule
        Baseline & 0.757 & 0.427 & -0.182 & -0.062 \\
        SDML$_{\textit{diag}}$ & 0.750 & 0.355 & -0.205 & -0.121 \\
        SDML$_{\mathrm{full}}$ & \textbf{0.774} & 0.404 & -0.122 & -0.037 \\ \bottomrule
    \end{tabular}
    \caption{Correlation analysis of model prediction with SCD task, polysemy (\#Synsets), and word frequency using Spearman's rank correlation on SemEval En.}
    \label{tab:ablation_relationship}
\end{table}

\section{Ablation Study}
\label{sec:ablation}
Motivated by the superior performance reported in \autoref{subsec:results_scd} by \ac{SDML}, we conduct an ablation study to understand the importance of (i) using sense-aware distance metric, and (ii) considering cross-dimensional correlations by Mahalanobis matrix.

% quantitative
Firstly, we conduct a quantitative comparison between our SDML (sense-aware sentence encoder and sense-aware distance metric using \emph{full} components of Mahalanobis matrix, SDML$_{\mathrm{full}}$) and two variants; (i) sense-aware sentence encoder only (Baseline), and (ii) sense-aware sentence encoder and sense-aware distance metric using \emph{diagonal} components of Mahalanobis matrix (SDML$_{\textit{diag}}$).
Results on SCD tasks are shown in \autoref{tab:ablation_scd}.
While our method achieves comparable or superior performance to the baseline even with the diagonal components of the Mahalanobis matrix (SDML$_{\textit{diag}}$), using the full Mahalanobis matrix (SDML$_{\mathrm{full}}$) yields further improvements. 
These results indicate the importance of considering inter-dimensional information by the full components of the sense-aware Mahalanobis matrix.

% qualitative
Secondly, we perform a qualitative analysis using gold labels.
Following \citet{aida-bollegala-2023-swap, aida-bollegala-2023-unsupervised}, we pick up the (i) top-8 semantically changed words, and (ii) top-8 semantically stable words in the SemEval En.
Additionally, due to the sense-aware method, we also evaluated the relationship between the SCD performance and two factors; polysemy and frequency as described in \citet{hamilton-etal-2016-diachronic}. 
We count the number of senses (synsets defined in WordNet) and frequency of the evaluation set of words in given corpora.
From \autoref{tab:ablation_analysis}, we can see that our method SDML$_{\mathrm{full}}$ slightly improves prediction.
Moreover, polysemous words tend to change their meanings (the law of innovation), but frequent words, contrary to the law of conformity, show no correlation with semantic change, as described in \cite{hamilton-etal-2016-diachronic}.
The same trends are in \autoref{tab:ablation_relationship}.
From this table, SDML$_{\textit{diag}}$ has a higher/lower correlation with frequency/polysemy than the baseline, which degrades the performance in SCD.
However, in SDML$_{\mathrm{full}}$, while a correlation with frequency/polysemy is slightly weaker than the baseline, they contribute to the performance improvement in SCD.
We conclude that relieving/enhancing the effect of frequency/polysemy will likely lead to further improvements in SCD performance.

% conclusion
Much prior work on \ac{SCD} use the Euclidean distance (or cosine similarity) for measuring semantic change scores of words~\cite{laicher-etal-2021-explaining,giulianelli-etal-2022-fire,rosin-etal-2022-time,rosin-radinsky-2022-temporal,cassotti-etal-2023-xl}, which (a) weights all dimensions equally, and (b) does not consider cross-dimensional correlations.
However, our experimental results show that Euclidean distance is a suboptimal choice for this purpose and learning a Mahalanobis distance metric is more appropriate.

\section{Conclusion}
We proposed a supervised two-staged \ac{SCD} method to address two challenges in the \ac{SCD} tasks: 1) models must obtain \emph{sense-aware} embeddings of target words over time, and 2) due to the lack of labelled training data, \ac{SCD} is often an unsupervised task using conventional distance metrics.
For the first challenge, we propose to learn a \emph{sense-aware} encoder.
Next, we address the second challenge by learning a \emph{sense-aware} distance metric to compare \emph{sense-aware} embeddings.
In both stages, we used \ac{WiC} datasets to provide sense-level supervision.
Experimental results show that our proposed method, \ac{SDML}, achieves strong performance %establishes a novel \ac{SoTA} 
in four \ac{SCD} benchmarks. %with 2-5\% improvements.
Moreover, \ac{SDML} achieves significant performance improvement in \ac{WiC} benchmarks.
Our findings highlight the importance of learning both a sense-aware encoder and a sense-aware distance metric.
%Moreover, we find that \emph{sense-aware} embeddings have \emph{\ac{SCD}-aware} dimensions, and our \emph{sense-aware} distance metric accurately exploits those dimensions for computing semantic change scores.

\section*{Limitations}
We show that our method can achieve strong %\ac{SoTA} 
performance in three languages (English, German, and Russian).
However, due to the limited training datasets, our method cannot perform well in other languages such as Swedish and Latin.
A potential solution to address this limitation and to further improve our method for languages not covered by existing WiC training datasets, is to explore the possibility of using cross-lingual language transfer methods.

\section*{Ethical Considerations}
In this paper, we focus on detecting semantic changes of words over time.
We did not create new datasets and used existing datasets for WiC and SCD for training and evaluation.
To the best of our knowledge, no ethical issues have been reported related to those datasets.
However, we used publicly available and pretrained \ac{MLM}s in this paper, and some of those \ac{MLM} are known to encode unfair social biases such as gender or race~\citep{basta-etal-2019-evaluating}.
It is possible that some of those social biases will be present (and possibly have been amplified) during the sense-aware encoder training process.
Therefore, we consider it to be an important and necessary task to evaluate the sense-aware encoder that we trained for any social biases before it is used in downstream tasks.

\bibliography{SCD.bib,anthology.bib}
\bibliographystyle{acl_natbib}

\appendix

\section*{Supplementary Materials}
\section{Details of \ac{ITML}}
\label{sec:appendix_ITML}

% add the pseudocode here and explain how u and l are set. Also mention about A_0

In this section, we describe the method we use to learn a sense-aware distance metric,
$h(\vec{w}_1, \vec{w}_2; \mat{A})$, that measures the semantic 
distance between two sense-aware embeddings $\vec{w}_1 = \vec{f}(w,s_1;\vec{\theta})$ and $\vec{w}_2 = \vec{f}(w,s_2;\vec{\theta})$, for $w$ in two sentences $s_1$ and $s_2$, respectively.
For the ease of reference, we re-write  \autoref{eq:metric} below as \autoref{app:eq:metric}, defining this distance metric.
\begin{align}
    \label{app:eq:metric}
    h(\vec{w}_1, \vec{w}_2; \mat{A}) = (\vec{w}_1 - \vec{w}_2)\T \mat{A}  (\vec{w}_1 - \vec{w}_2)
\end{align}
Here, $\mat{A} \in \R^{d \times d}$ is a Mahalanobis matrix (assumed to be positive definite) that defines a (squared) distance metric.

To learn the Mahalanobis distance matrix $\mat{A}$ in \autoref{app:eq:metric}, we use training instances $(w, s_1, s_2, y)$ from a \ac{WiC} dataset for a target word $w$ and two sentences $s_1$ and $s_2$, where $y=1$ indicates that $w$ has the same meaning in both $s_1$ and $s_2$, whereas $y=0$ indicates that $w$ takes different meanings.

We consider two types of constraints that must be satisfied by $h$ as follows.
For instances where $y=1$, the distance between $\vec{w}_1$ and $\vec{w}_2$ is required to be greater than a distance lower bound $\ell$ such that, $h(\vec{w}_1, \vec{w}_2) \geq \ell$.
Likewise, for instances where $y=0$, the distance between $\vec{w}_1$ and $\vec{w}_2$ is required to be lower than a distance upper bound $u$ such that, $h(\vec{w}_1, \vec{w}_2) \leq u$.
Recall that there exists a bijection (up to a scaling function) between the set of Mahalanobis distances and the set of the equal mean (denoted by $\vec{\mu}$) multivariate Gaussian distributions given by $p(\vec{w}; \mat{A}) = \frac{1}{Z} \exp\left(-\frac{1}{2} h(\vec{w}, \vec{\mu}; \mat{A})\right)$, where $Z$ is a normalising constant and $\mat{A}\inv$ is the covariance of the distribution.
We can use this bijection to measure the distance between two Mahalanobis distance functions parametrised by $\mat{A}_0$ and $\mat{A}$ using the relative entropy ( KL-divergence) between the corresponding multivariate Gaussians, given by \autoref{app:eq:KL}.
\begin{align}
\small
    \label{app:eq:KL}
    \mathrm{KL}(p(\vec{w}; \mat{A}_0) \vert\vert p(\vec{w}; \mat{A})) = \int p(\vec{w}; \mat{A}_0) \log \frac{p(\vec{w}; \mat{A}_0)}{p(\vec{w}; \mat{A})}
\end{align}
Relative entropy is a non-negative and convex function in $\mat{A}$.
The overall optimisation problem is given by \autoref{app:eq:opt}.
\begin{align}
    \label{app:eq:opt}
		\min_{\mat{A}}  \quad & \mathrm{KL}(p(\vec{w}; \mat{A}_0) \vert\vert p(\vec{w}; \mat{A})) \nonumber \\
    \text{subject to} \quad & h(\vec{w}_1, \vec{w}_2) \leq u  \quad y=1, \\ \nonumber
                    & h(\vec{w}_1, \vec{w}_2) \geq \ell \quad y=0 .
\end{align}

The optimisation problem given in \autoref{app:eq:opt} can be expressed as a particular type of Bregman divergence, which can be efficiently solved using the Bregman's method~\citep{Censor:1997}.
We use the \ac{ITML} algorithm proposed by \citet{davis-etal-2007-itml} to learn $\mat{A}$, which is described in \autoref{algo:ITML}.
We arrange the sense-aware representations $\vec{f}(w,s_1)$ and $\vec{f}(w, s_2)$ for each instance $(w,s_1,s_2,y)$ as columns to create the input matrix $\mat{X} \in R^{d \times 2n}$, where $d$ is the dimensionality of the sense-aware embeddings produced by the encoder trained in \autoref{sec:encoder}, and $n$ is the total number of training instances in the \ac{WiC} dataset.
The initial value of the distance matrix, $\mat{A}_0$ is set to the identity matrix, which corresponds to computing the Euclidean distance.
The upper bound, $u$, is set to the distance that covers the top $5\%$ of the distances between positive instances (i.e. $y=1$), while the lower bound, $l$, is set to the distance that covers the bottom $5\%$ of the distances between the negative instances ($y=0$).

\begin{algorithm}[t]
    \small
    \algsetup{linenosize=\small}
    \caption{Information Theoretic Metric Learning (ITML)}
    \label{algo:ITML}
    \begin{algorithmic}[1]  % remove [1] to hide the line numbers.
    \renewcommand{\algorithmicrequire}{\textbf{Input:}}
    \renewcommand{\algorithmicensure}{\textbf{Output:}}
    \REQUIRE input matrix $\mat{X}$, labels $\vec{y}$, distance thresholds $[l, u]$, \\
    input Mahalanobis matrix $\mat{A}_0$, slack parameter $\gamma$
    \ENSURE Mahalanobis matrix $\mat{A}$
    \STATE \texttt{\# Initialise $\mat{A}$, $\vec{\lambda}$, and $\vec{\xi}$} 
    \STATE $\mat{A} \leftarrow \mat{A}_0$
    \FOR{$i=1$ to $n$}
        \STATE $\lambda_i \leftarrow 0$
        \STATE $\xi_i \leftarrow u$ \textbf{if} $y_i = 1$ \textbf{else} $l$
    \ENDFOR
    \STATE \texttt{\# Optimise $\mat{A}$}
    \REPEAT
        \FOR{$i=1$ to $n$}
            \STATE obtain $i$-th instance $(\vec{w}_1, \vec{w}_2, y_i)$ from $\mat{X}$ and $\vec{y}$
            \STATE $d \leftarrow h(\vec{w}_1, \vec{w}_2; \mat{A})$ in \autoref{app:eq:metric}
            \STATE $\delta \leftarrow 1$ \textbf{if} $y_i = 1$ \textbf{else} $-1$
            %\STATE $\alpha \leftarrow \mathrm{min}(\lambda_i, \frac{\delta}{2}(\frac{1}{d} - \frac{\gamma}{\xi_i}))$
            \STATE $\alpha \leftarrow \mathrm{min}(\lambda_i, \delta(1/d - \gamma/\xi_i)/2)$
            \STATE $\xi_i \leftarrow \gamma \xi_i / (\gamma + \delta\alpha \xi_i)$
            \STATE $\lambda_i \leftarrow \lambda_i - \alpha$
            \STATE $\beta \leftarrow \delta\alpha/(1 - \delta\alpha d)$
            \STATE $\mat{A} \leftarrow \mat{A} + \beta\alpha(\vec{w}_1 - \vec{w}_2)(\vec{w}_1 - \vec{w}_2)\T \mat{A}$
        \ENDFOR
    \UNTIL convergence
    \RETURN $\mat{A}$
    %\STATE $\cS^{+}_{\rm def}, \cS^{-}_{\rm def} \leftarrow \mathrm{RandomSplit}(\cS_{\rm def}, t)$
    %\STATE $\cS^{*} \leftarrow \emptyset$
    %\STATE $~\alpha~\leftarrow 0$
    %\FOR {$\cS \in (\cS^{-}_{\rm def} \cup \cS_{\rm div})$}
    %    \IF {$(|\cS| == N) \land (f(\cS) \geq \alpha)$}
    %        \STATE $\alpha \leftarrow f(\cS)$
    %        \STATE $\cS^* \leftarrow \cS$
    %    \ENDIF
    %\ENDFOR
    %\RETURN $\cS^{*}$
    \end{algorithmic}
\end{algorithm}

\section{Data Statistics}
\label{sec:appendix_data}
We provide the full data statistics for the \ac{SCD} benchmarks in \autoref{tab:result_scd}.

%\begin{table}[t]
\begin{table*}[h]
    \centering
    \small{
    \begin{tabular}{llccrrr} \toprule
        Dataset & Language & Time Period & \#Targets & \#Sentences & \#Tokens & \#Types \\ \midrule
        \multirow{8}{*}{SemEval-2020 Task 1} & \multirow{2}{*}{English} & 1810--1860 & \multirow{2}{*}{37} & 254k & 6.5M & 87k  \\
        %\multirow{8}{*}{SemEval} & \multirow{2}{*}{English} & 1810--1860 & \multirow{2}{*}{37} & 254k \\
         & & 1960--2010 & & 354k & 6.7M & 150k \\
         & \multirow{2}{*}{German} & 1800--1899 & \multirow{2}{*}{48} & 2.6M & 70.2M & 1.0M \\
         & & 1946--1990 & & 3.5M & 72.3M & 2.3M \\
         & \multirow{2}{*}{Swedish} & 1790--1830 & \multirow{2}{*}{31} & 3.4M & 71.0M & 1.9M \\
         & & 1895--1903 & & 5.2M & 110.0M & 3.4M \\
         & \multirow{2}{*}{Latin} & B.C. 200--0 & \multirow{2}{*}{40} & 96k & 1.7M & 65k  \\
         & & 0--2000 & & 463k & 9.4M & 253k  \\ \midrule
        \multirow{3}{*}{RuShiftEval} & \multirow{3}{*}{Russian} & 1700--1916 & \multirow{3}{*}{99} & 3.3k & 97k & 39k \\
         & & 1918--1990 & & 3.3k & 78k & 34k \\
         & & 1992--2016 & & 3.3k & 78k & 35k \\ \bottomrule %\midrule
        %\multirow{2}{*}{RuShiftEval1} & \multirow{2}{*}{Russian} & 1700--1916 & \multirow{2}{*}{99} & 3.3k & 97k & 39k \\
        % & & 1918--1990 & & 3.3k & 78k & 34k \\
        %\multirow{2}{*}{RuShiftEval2} & \multirow{2}{*}{Russian} & 1918--1990 & \multirow{2}{*}{99} & 3.3k & 78k & 34k \\
        % & & 1992--2016 & & 3.3k & 78k & 35k \\
        % \multirow{2}{*}{RuShiftEval3} & \multirow{2}{*}{Russian} & 1700--1916 & \multirow{2}{*}{99} & 3.3k & 95k & 38k \\
        % & & 1992--2016 & & 3.3k & 78k & 35k \\ \midrule
         % \multirow{2}{*}{LiverpoolFC} & \multirow{2}{*}{English} & 2011--2013 & \multirow{2}{*}{97} & 576k & 9.5M & 137k \\
         %& & 2017 & & 1.0M & 15.7M & 146k \\\bottomrule
    \end{tabular}
    }
    \caption{Full statistics of the SCD datasets.}
    \label{tab:data_scd_full}
%\end{table}
\end{table*}

\end{document}